%% file: main.tex
\documentclass[sigconf]{acmart}
\usepackage{multirow}
\usepackage{booktabs}
\usepackage{pifont}
\usepackage{xcolor}
\usepackage{soul}
\usepackage[normalem]{ulem}
\usepackage{balance}

\definecolor{darkred}{rgb}{0.39,0.04,0.45}
\AtBeginDocument{%
  }

\copyrightyear{2026}
\acmYear{2026}
\setcopyright{cc}
\setcctype{by}
\acmConference[MM '26]{Proceedings of the 34th ACM International Conference on Multimedia}{November 10--14, 2026}{Rio de Janeiro, Brazil}
\acmBooktitle{Proceedings of the 34th ACM International Conference on Multimedia (MM '26), November 10--14, 2026, Rio de Janeiro, Brazil}
\acmDOI{10.1145/3767308.3835263}
\acmISBN{979-8-4007-2213-4/2026/11}

\begin{document}

\title{GRID-Planner: Graph-Structured Relational Reward Shaping for Diffusion Planning}

\author{Zhao Yang}
\email{lingxi.yz950701@gmail.com}
\affiliation{%
  \institution{Xi'an Jiaotong University}
  \city{Xi'an}
  \country{China}
}

\author{Zezhong Qian}
\email{zezhongqian@stu.xjtu.edu.cn}
\affiliation{%
  \institution{Xi'an Jiaotong University}
  \city{Xi'an}
  \country{China}
}

\author{Yinan Shi}
\email{syn1211@outlook.com}
\affiliation{%
  \institution{Cytoderm Intelligent Technology}
  \city{Beijing}
  \country{China}
}

\author{Longjun Liu}
\correspondingauthor
\email{liulongjun@xjtu.edu.cn}
\affiliation{%
  \institution{Xi'an Jiaotong University}
  \city{Xi'an}
  \country{China}
}

\author{Yawei Jueluo}
\email{jueluo.yawei@cybopal.com}
\affiliation{%
  \institution{Cytoderm Intelligent Technology}
  \city{Beijing}
  \country{China}
}

\renewcommand{\shortauthors}{Zhao Yang, Zezhong Qian, Yinan Shi, Longjun Liu, and Yawei Jueluo}
\begin{abstract}
Autonomous driving planning requires selecting trajectories that remain safe, rule-compliant, and comfortable under complex multi-agent interactions, yet robust closed-loop decision making in planning-critical scenes remains challenging. Existing diffusion- and RL-based planners often rely on generic rewards and use relational structure mainly for representation, limiting how planning-critical rollouts shape policy improvement. We present GRID-Planner, a graph--reward integrated diffusion planning framework that aligns generative planning with structured relational signals. Specifically, GRID-Planner combines graph-aware importance weighting, a time-indexed relational reward surrogate, and an advantage-weighted diffusion objective with on-policy self-distillation, forming a rollout-return-advantage-update loop without explicit likelihoods. Experiments on nuPlan and interPlan show that, compared with the diffusion-only baseline, GRID-Planner reduces collision rate, lane deviation, and jerk by 35.8\%, 60.8\%, and 27.4\%, respectively, and improves the planning score by 3.4\%. \textbf{Project website:} \url{https://github.com/yangzhaojason/GRID-Planner}.
\end{abstract}



\maketitle

\input{sec/1_intro_final}
\input{sec/2_related_final}
\input{sec/3_method}
\input{sec/4_exp}

\section{Conclusion}
\label{sec:conclusion}
We presented \textbf{GRID-Planner}, a graph--reward integrated diffusion planner that aligns generative planning with structured relational signals for autonomous driving. By combining \textbf{Graph-Aware Importance Weighting} (GAIW), a \textbf{Relational Reward Surrogate} (RRS), and a unified on-policy diffusion objective with actor--critic and self-distillation, our framework improves trajectory prioritization, rollout evaluation, and policy refinement in a coherent manner. \noindent\textbf{Limitations and Future Work.} Experiments on \textbf{nuPlan} and \textbf{interPlan} show that \textbf{GRID-Planner} consistently improves closed-loop performance, especially in reactive and interaction-intensive scenarios, while our current design still relies on hand-crafted graph weighting and a limited set of relational reward signals; future work will explore richer relational cues, more adaptive graph-based weighting, and more efficient on-policy optimization for large-scale real-world autonomous driving deployment.

\section*{Acknowledgment} 
This work was supported by the Fundamental and Interdisciplinary Disciplines Breakthrough Plan of the Ministry of Education of China under Grant No.~JYB2025XDXM504; the National Natural Science Foundation of China under Grant Nos.~62573343 and 62088102; the National Basic Strengthen Research Program of ReRAM under Grant No.~2022-00-03; the Fundamental Research Funds for the Central Universities under Grant No.~xzy012024066; and the Open-End Fund of Beijing Institute of Control Engineering under Grant No.~OBCandETL-2024-04.

\bibliographystyle{ACM-Reference-Format}
\balance
\bibliography{references}

\end{document}

%% file: sec/1_intro_final.tex
\section{Introduction}
\label{sec:intro}

Autonomous driving planning requires selecting ego trajectories that are safe, rule-compliant, and comfortable under complex multi-agent interactions~\cite{caesar2021nuplan,hu2023planning,dauner2023misconceptions,jia2023driveadapter,bench2drive2024}. Despite rapid progress in learning-based planning, robust closed-loop decision making in dense urban scenes remains challenging~\cite{caesar2021nuplan,dauner2023misconceptions,bench2drive2024}. A core challenge is that planning performance depends not only on average-case trajectory realism but on learning from a small set of \emph{planning-critical} trajectories---those arising in highly interactive, safety-sensitive, and rule-constrained scenarios such as merging, yielding, and intersection conflicts~\cite{caesar2021nuplan,dauner2023misconceptions,bench2drive2024}. In practice, however, most training trajectories are semantically flat, weakly constrained, and low-conflict, contributing limited signals to the decision boundaries that matter most for planning~\cite{dauner2023misconceptions}. This long-tail structure makes pure imitation learning insufficient, since abundant easy trajectories dominate supervision, while naive reinforcement learning is also inadequate because generic rewards are difficult to design stably for complex multi-agent interactions. These challenges motivate graph-structured relational modeling to prioritize planning-critical trajectories and support planning-oriented policy improvement.

\begin{figure}[t]
    \centering
    \includegraphics[width=0.47\textwidth]{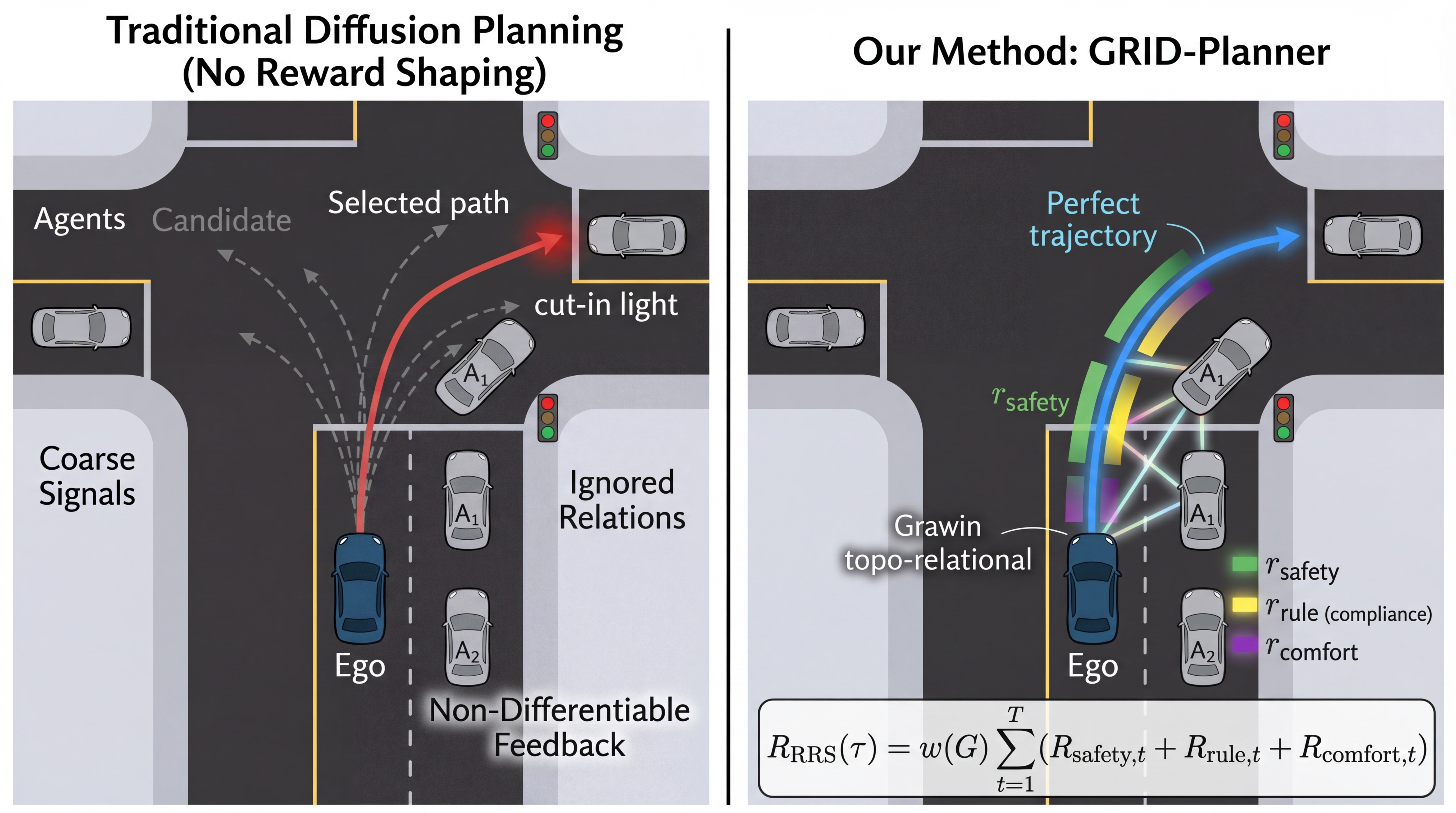}
    \caption{\textbf{Overview of GRID-Planner.} Traditional diffusion planning may miss planning-critical relational semantics, whereas GRID-Planner combines Graph-Aware Importance Weighting and RRS to prioritize high-risk trajectories and evaluate safety, lane/route consistency, and comfort.}
    \label{fig:intro_motivation}
    \vspace{-0.4cm}
\end{figure}

Existing methods improve planning from different angles, but the core issue remains unresolved. Strong imitation-based planners, such as UrbanDriver~\cite{scheel2022urbandriver}, PlanTF~\cite{cheng2023plantf}, and PLUTO~\cite{cheng2024pluto}, optimize similarity to expert behavior, allowing abundant easy trajectories to dominate training and under-emphasizing safety-critical and rule-sensitive cases. Interaction-centric methods, such as GameFormer~\cite{huang2023gameformer}, improve multi-agent relation modeling, but mainly at the representation level, without explicitly using relational structure to determine trajectory value or optimization priority. Diffusion-based planners, such as Diffusion Planner~\cite{zheng2025diffusionplanner}, provide flexible multi-modal trajectory generation, while reward-guided variants such as Diffusion-ES~\cite{yang2024diffusiones} demonstrate the promise of reward-driven optimization. However, existing approaches still rely either on behavior cloning objectives, which suffer from compounding errors and exposure bias in closed-loop inference, or on generic reward-driven optimization, which does not explicitly encode planning-oriented relational semantics. This leaves open how relational semantics should shape both the training distribution and the optimization objective, rather than serving only as scene representations or sampling-time guidance.

To address this gap, we propose \textbf{GRID-Planner}, a \textbf{G}raph-\textbf{R}eward \textbf{I}ntegrated \textbf{D}iffusion planner that uses graph-structured scene semantics to directly shape the optimization signal of generative planning. Rather than treating graph structure as just an input representation, we use it to define both trajectory importance and planning-oriented return. Our framework answers three questions left open by prior planners: which trajectories should matter more during training, how their quality should be evaluated for planning, and how relational signals should reshape diffusion-based policy learning. Specifically, our framework contains three tightly coupled components: \textcolor{darkred}{\ding{182}} \textbf{Graph-Aware Importance Weighting}, which upweights trajectories from complex, high-risk scenes using scene-interaction centrality and collision-risk cues; \textcolor{darkred}{\ding{183}} \textbf{Relational Reward Surrogate} (RRS), which evaluates candidate rollouts through time-indexed ego-centric scene graphs, producing a dense trajectory-level signal with smooth bounded penalties over safety, lane/route consistency, and comfort; and \textcolor{darkred}{\ding{184}} \textbf{Graph-Reward Integrated Diffusion Objective}, which incorporates structured returns into a low-variance advantage-weighted diffusion objective, coupled with an advantage-gated on-policy self-distillation (OPSD) mechanism. The resulting advantage reweights diffusion denoising and gates OPSD teacher selection and updates, forming a rollout--return--advantage--update loop in which graph-reward signals directly drive policy improvement.

The main contributions of this work are:
\begin{itemize}
    \item We propose \textbf{GRID-Planner}, a unified graph-reward integrated diffusion framework that redefines how scene graphs are used in autonomous driving planning: rather than serving only as input representations, they directly shape optimization priority and trajectory value for closed-loop decision making.
    \item We introduce a \textbf{planning-oriented relational learning mechanism}, instantiated by \textbf{Graph-Aware Importance Weighting} and a \textbf{Relational Reward Surrogate}, which together focus training on planning-critical trajectories and score candidate rollouts using safety, lane/route consistency, and comfort criteria.
    \item We develop a \textbf{unified on-policy diffusion objective} that combines advantage-weighted diffusion with advantage-gated self-distillation to support stable on-policy refinement and consolidate high-return behaviors, yielding consistent gains over strong baselines on nuPlan and interPlan.
\end{itemize}

\begin{figure*}[t]
    \centering
    \includegraphics[width=\textwidth]{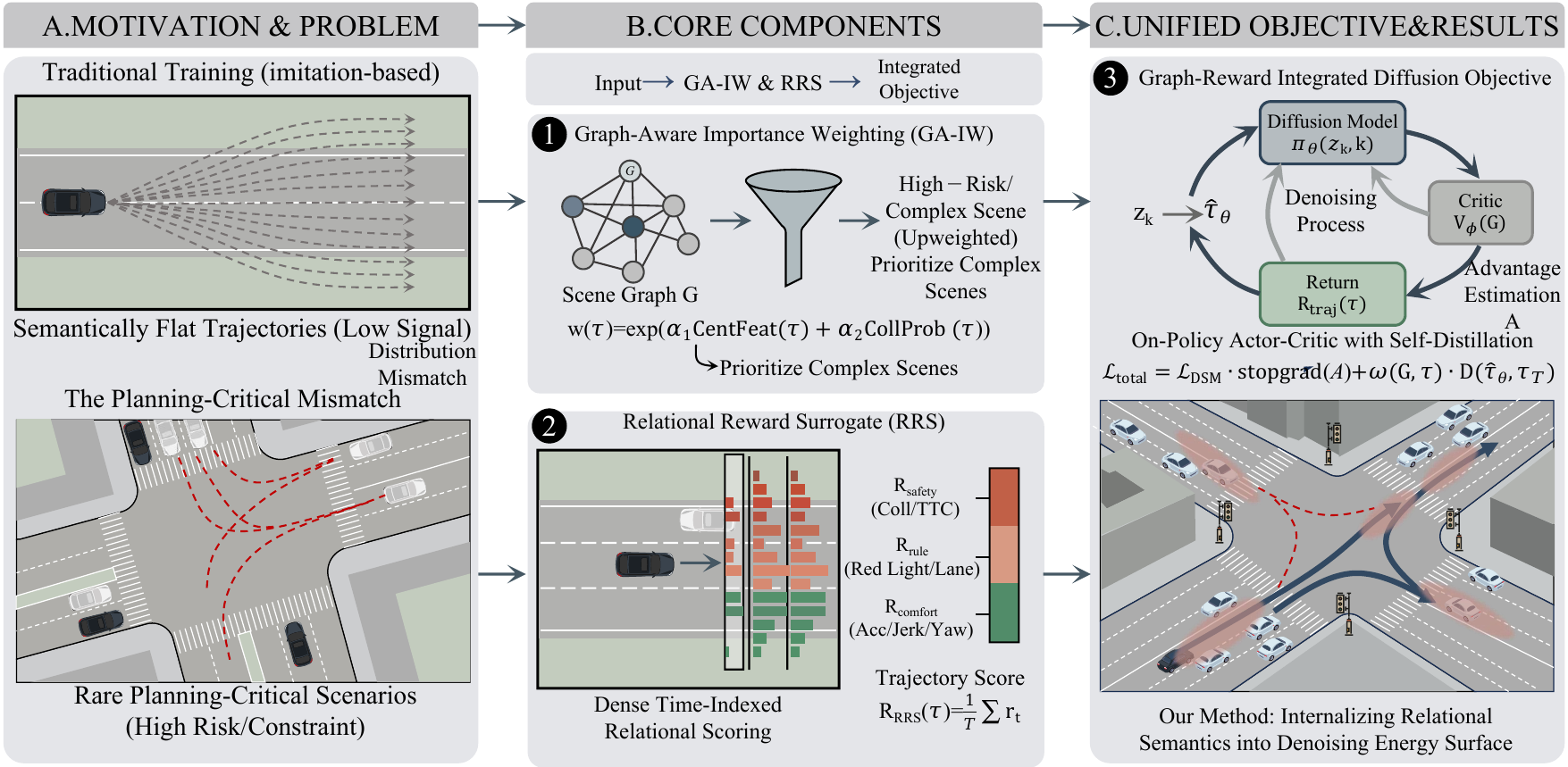}
    \caption{Overall framework of GRID-Planner. We address the mismatch between semantically flat imitation signals and rare planning-critical scenarios through two components, graph-aware importance weighting (GA-IW) and relational reward surrogate (RRS), which are unified in the final graph-reward integrated diffusion objective.}
    \label{fig:grid_framework}
    \vspace{-0.2cm}
\end{figure*}

%% file: sec/2_related_final.tex
\section{Related Work}
\label{sec:related}

\noindent\textbf{Imitation-Based and Interaction-Centric Planning.}
A major line of work learns autonomous driving policies from expert demonstrations~\cite{caesar2021nuplan,dauner2023misconceptions,hu2023planning,prakash2021transfuser,wu2022tcp,jia2023thinktwice,jia2023driveadapter}. UrbanDriver~\cite{scheel2022urbandriver}, PlanTF~\cite{cheng2023plantf}, and PLUTO~\cite{cheng2024pluto} achieve strong closed-loop performance through carefully designed imitation objectives, while UniAD~\cite{hu2023planning}, VAD~\cite{jiang2023vad},
VADv2~\cite{chen2024vadv2}, SparseDrive~\cite{sun2024sparsedrive},
DTPP~\cite{huang2023dtpp}, GameFormer~\cite{huang2023gameformer},
InterFuser~\cite{shao2023interfuser}, and Causal-Planner~\cite{yuan2025causalplanner}
improve scene representation and interaction modeling. However, these approaches mainly optimize expert imitation or relational representations, without explicitly using scene relations to determine trajectory importance or planner-oriented value. GRID-Planner instead uses graph structure to prioritize planning-critical trajectories and construct relational returns over safety, lane/route consistency, and comfort.

\noindent\textbf{Generative Planning with Diffusion.}
Generative planners model future trajectories as samples from expressive distributions, supporting multi-modal driving behaviors~\cite{jiang2023motiondiffuser,wang2024drivewm}. GenAD~\cite{zheng2024genad}, Diffusion Planner~\cite{zheng2025diffusionplanner}, and Diffusion-ES~\cite{yang2024diffusiones} explore latent generation, diffusion-based closed-loop planning, and reward-guided sampling, respectively. Flow Planner~\cite{tan2025flow}, DiffusionDrive~\cite{liao2025diffusiondrive}, and BridgeDrive~\cite{liu2026bridgedrive} further improve interactive modeling and diffusion-policy design. These methods primarily focus on trajectory generation or sampling-time guidance, whereas GRID-Planner incorporates structured relational signals into training through trajectory prioritization, relational evaluation, and diffusion-policy refinement.

\noindent\textbf{RL-Based Policy Improvement for Generative Planning.}
Reinforcement learning has been combined with diffusion policies for reward-driven optimization. DDPO~\cite{black2024trainingdiffusionmodelsreinforcement}, advantage-weighted regression methods~\cite{peng2019advantageweightedregressionsimplescalable,nair2021awacacceleratingonlinereinforcement}, and diffusion policies~\cite{wang2023diffusionpoliciesexpressivepolicy} adapt policy-improvement principles to expressive generative models. However, they generally rely on generic scalar rewards rather than planning-oriented relational semantics. GRID-Planner couples graph-weighted relational returns with advantage-weighted denoising and advantage-gated on-policy self-distillation, allowing relational signals to directly shape policy updates.

%% file: sec/3_method.tex
\section{Methodology}
\label{sec:method}
\subsection{Preliminaries}
\label{ssec:prelim}

\noindent\textbf{Notation.}
We denote an ego candidate trajectory by $\tau=\{(s_t, a_t)\}_{t=1}^T$
over horizon steps $t\in\{1,\dots,T\}$, where $s_t$ is the ego kinematic state (position $p^{\text{ego}}_t\in\mathbb R^2$ or $\mathbb R^3$, velocity, and heading), and $a_t$ is the control action in an ego-centric frame. Neighboring trajectories are $\{s^{(j)}_{1:T}\}_{j=1}^N$ with neighbor index $j\in\{1,\dots,N\}$.
We use $t=0$ to denote the current step (the last observed frame), with positions $p^{\text{ego}}_0$ and $p^{(j)}_0$. We define $\mathrm{clip}(x,0,1)=\min(\max(x,0),1)$.

\noindent\textbf{Diffusion trajectory model and energy shaping.}
We model the trajectory distribution $\pi_\theta(\tau)$ using a score-based diffusion process. Given a data trajectory $\tau_{\mathrm{data}}$, the forward noising process at diffusion step $k \in (0, 1]$ is:
\begin{equation}
z_k=\sqrt{\bar\alpha_k}\,\tau_{\mathrm{data}}+\sqrt{1-\bar\alpha_k}\,\epsilon,
\qquad \epsilon\sim\mathcal N(0,I),
\end{equation}
where the model predicts the score $\epsilon_\theta(z_k,k)=-\sigma_k s_\theta(z_k,k)$ with $\sigma_k^2=1-\bar\alpha_k$, trained by
\begin{equation}
\mathcal L_{\mathrm{DSM}} = \mathbb E_{k,\tau_{\mathrm{data}},\epsilon}\left[\|\epsilon - \epsilon_\theta(z_k,k)\|^2\right].
\end{equation}
Sampling proceeds by running the reverse denoising dynamics from $z_1 \sim \mathcal{N}(0,I)$, yielding $\tau \sim \pi_\theta$. Under maximum-entropy RL, the optimal trajectory distribution can be expressed as
$\pi^\star(\tau) \propto \exp(Q^\star(\tau)/\lambda)$, motivating us to shape $\pi_\theta(\tau)$ using trajectory energies derived from surrogate return functions.

\subsection{Problem Statement}
\label{ssec:problem}

At each planning step, we represent the driving scene as a relational graph $G=(V,E)$, where nodes $V$ include the ego vehicle, surrounding agents, lane polylines, and traffic signals, and edges $E$ encode geometric relations and traffic-rule priorities. 
We consider trajectory planning over a candidate trajectory $\tau$ and define the trajectory-level relational score
\begin{equation}
\mathcal R_{\mathrm{RRS}}(\tau) = R_{\mathrm{safety}}(\tau) + R_{\mathrm{rule}}(\tau) + R_{\mathrm{comfort}}(\tau),
\label{eq:RRS}
\end{equation}
which evaluates safety, rule compliance, and ride comfort (each term is specified in the next subsection).

To align this trajectory-level evaluator with entropy-regularized distribution shaping, we use a trajectory return surrogate:
\begin{equation}
R_{\mathrm{traj}}(\tau) = w(\tau) \mathcal R_{\mathrm{RRS}}(\tau),
\label{eq:Qsoft_general}
\end{equation}
where $w(\tau) \ge 0$ is an importance weight used to differentiate candidate rollouts. Our goal is to learn a graph-conditioned trajectory distribution $\pi_\theta(\tau|G)$ that concentrates probability mass on trajectories with higher $R_{\mathrm{traj}}(\tau)$, while maintaining diversity and efficient inference. To achieve this, we employ a unified graph-reward integrated diffusion objective (Sec.~\ref{ssec:grid_obj}), which combines advantage-weighted diffusion updates with a state-value baseline and on-policy self-distillation, aligning training with planner-induced rollouts and consolidating high-reward behaviors, as illustrated in Fig.~\ref{fig:grid_framework}.
\subsection{Graph-Aware Weighting for Semantic Alignment}
\label{ssec:gaiw}

\begin{figure}[t]
    \centering
    \includegraphics[width=0.47\textwidth]{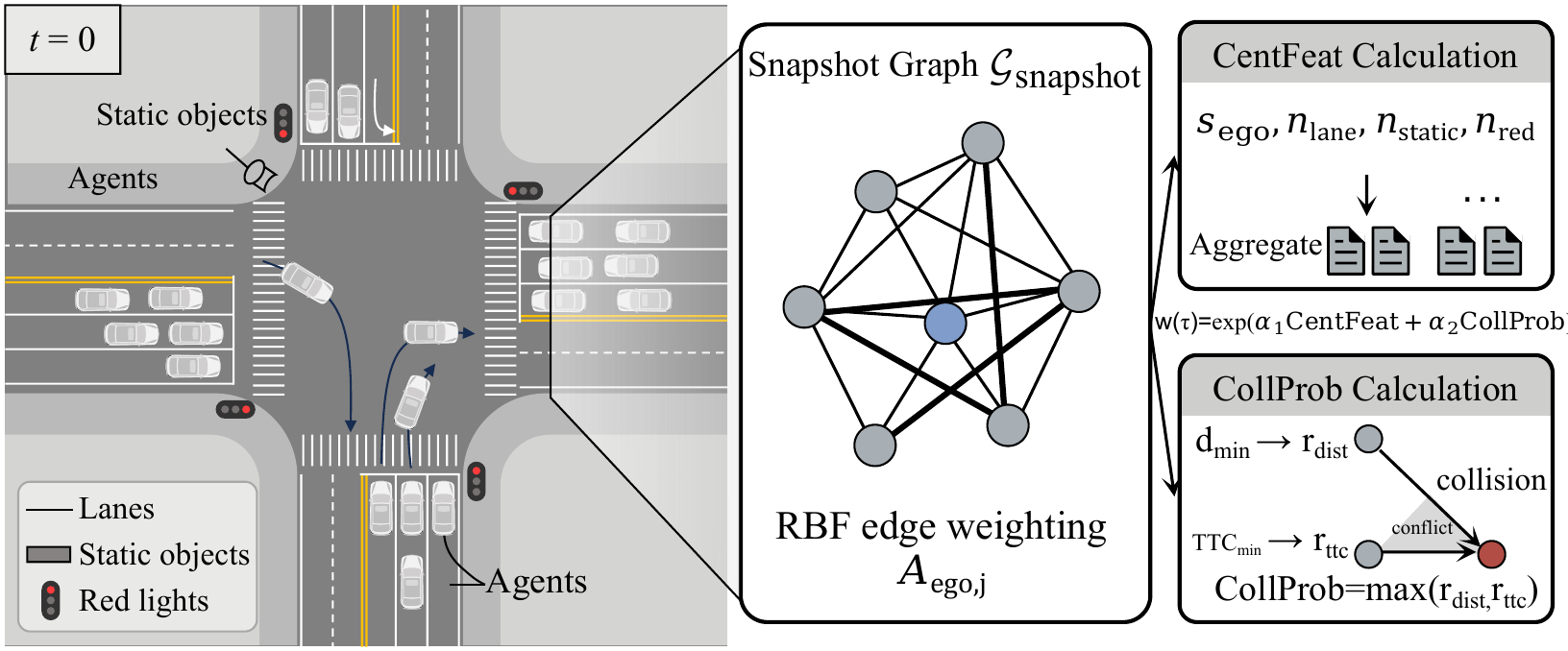}
    \caption{Illustration of the graph-aware weighting mechanism. $\mathrm{CentFeat}$ is computed from a snapshot interaction graph at $t=0$ and map-element counts, while $\mathrm{CollProb}$ is estimated from future proximity and TTC. They are combined to form the importance weight $w(\tau)$ in Eq.~\eqref{eq:gaiw_weight}.}
    \label{fig:gaiw_overview}
    \vspace{-0.6cm}
\end{figure}

Real-world driving scenarios are highly imbalanced, with most trajectories being semantically flat and only a small fraction being planning-critical: most trajectories are
``flat'' (neither risky nor strongly constrained), while only a small fraction is
safety-critical. To prevent gradients from being dominated by abundant
"flat" samples, we introduce a \emph{graph-aware importance weight}
$w(\tau)$ that emphasizes structurally complex and high-risk trajectories.
The ``graph'' here is a lightweight \emph{scene-interaction proxy} used only
for weighting (distinct from the model conditioning graph), as illustrated in
Fig.~\ref{fig:gaiw_overview}.

\noindent\textbf{Weight definition.}
For each candidate ego trajectory $\tau=\{(s_t, a_t)\}_{t=1}^T$, we define
\begin{equation}
w(\tau)=\exp\!\big(\alpha_1\,\mathrm{CentFeat}(\tau)+\alpha_2\,\mathrm{CollProb}(\tau)\big),
\label{eq:gaiw_weight}
\end{equation}
which yields the trajectory-level return
$R_{\mathrm{traj}}(\tau)=w(\tau)\,\mathcal R_{\mathrm{RRS}}(\tau)$.
Here $\alpha_1,\alpha_2$ are scalar weights.
$\mathrm{CentFeat}(\tau)$ measures local interaction and map complexity at the current step ($t=0$) to prioritize complex scenes, while $\mathrm{CollProb}(\tau)\in[0,1]$
is a bounded collision-risk score that combines future proximity along the rollout and current time-to-collision (TTC) computed at $t=0$. Although written as $\mathrm{CentFeat}(\tau)$ for notational simplicity, it depends only on the current snapshot scene at $t=0$. Notably, $w(\tau)$ serves as a training-time factor rather than a reward bonus for risk-taking.

\noindent\textbf{Past-snapshot interaction graph and centrality.}
We construct a snapshot interaction graph at $t=0$ from the last observed frame, denoted by $\mathcal G_{\text{snapshot}}$. Its nodes are $\{\text{ego},1,\dots,N\}$, with ego-frame positions $p^{\text{ego}}_0$ and $p^{(j)}_0$. We connect ego--agent pairs within a radius $R_g$ and assign an RBF edge weight
\begin{equation}
A_{\text{ego},j}=\exp\!\left(-\frac{\|p^{\text{ego}}_0-p^{(j)}_0\|^2}{2\sigma_{\mathrm{rbf}}^2}\right)\cdot
\mathbf{1}\!\left[\|p^{\text{ego}}_0-p^{(j)}_0\|\le R_g\right],
\end{equation}
where $\sigma_{\mathrm{rbf}}$ is the kernel bandwidth and $\mathbf{1}[\cdot]$ is the indicator.
The ego interaction strength (weighted degree) is
$s_{\text{ego}}=\sum_{j=1}^{N}A_{\text{ego},j}$.
To reflect local map complexity, we count nearby map elements around the ego at $t=0$
(with fixed radii): the number of lane polylines $n_{\text{lane}}$, static objects
$n_{\text{static}}$, and red-light signal points $n_{\text{red}}$. The final centrality
feature is
\begin{equation}
\begin{aligned}
\mathrm{CentFeat}(\tau) \;=\;& \log(1+s_{\text{ego}})
+ \beta_{\text{lane}}\log(1+n_{\text{lane}}) \\
&+ \beta_{\text{static}}\log(1+n_{\text{static}})
+ \beta_{\text{red}}\log(1+n_{\text{red}}),
\end{aligned}
\label{eq:centfeat}
\end{equation}
where $\beta_{\text{lane}},\beta_{\text{static}},\beta_{\text{red}}$ are scalar coefficients.

\noindent\textbf{Collision-risk proxy.}
We estimate $\mathrm{CollProb}(\tau)$ from (i) future proximity over the horizon and
(ii) current TTC at $t=0$, where TTC is computed from relative position and velocity. To avoid numerical instability and singularities, we convert the current TTC into a bounded inverse-TTC (iTTC) risk metric.
For future proximity, over neighbors $j\in\{1,\dots,N\}$ and horizon steps $t\in\{1,\dots,T\}$,
we compute
\begin{equation}
\begin{aligned}
d_{\min} &= \min_{j,t}\ \bigl\|p^{(j)}_t - p^{\text{ego}}_t\bigr\|,\\
r_{\text{dist}} &=
\mathrm{clip}\!\left(\frac{R_{\text{coll}}-d_{\min}}{R_{\text{coll}}},\,0,\,1\right),
\end{aligned}
\label{eq:risk_dist}
\end{equation}
where $p^{\text{ego}}_t$ is the ego position from $\tau$, while $p^{(j)}_t$ uses log-replay futures offline and constant-velocity extrapolation on-policy; the same sources are used by RRS safety, and no neighbor futures are queried at inference. $R_{\text{coll}}$ is a collision-distance threshold, and
$\mathrm{clip}(\cdot,0,1)$ bounds the score to $[0,1]$.

For the current TTC risk at $t=0$, let $v^{\mathrm{rel}}_{0,j}$ denote the line-of-sight relative speed of neighbor $j$. We compute the risk proxy $r_{\text{ttc}}$ via
\begin{equation}
\begin{aligned}
\mathrm{iTTC}_{\max} &= \max_{j}\ \mathrm{clip}\!\left(\frac{\max(0, -v^{\mathrm{rel}}_{0,j})}{\max(\varepsilon, \|p^{\text{ego}}_0 - p^{(j)}_0\|)},\, 0,\, \mathrm{iTTC}_{\text{cap}}\right),\\
r_{\text{ttc}} &= 1 - \exp\!\left(-\mathrm{iTTC}_{\max} \cdot s_{\text{ttc}}\right),\\
\mathrm{CollProb}(\tau) &= \max\!\left(r_{\text{dist}},\,r_{\text{ttc}}\right),
\end{aligned}
\label{eq:collprob}
\end{equation}
where $s_{\text{ttc}}>0$ controls the decay scale and $\mathrm{iTTC}_{\text{cap}}$ is an upper bound.
In practice, we stabilize training by clipping $w(\tau)$ to $[w_{\min},w_{\max}]$
and normalizing it by the batch mean.
\subsection{Relational Reward Surrogate}
\label{ssec:rrs}
\begin{figure}[t]
\centering
\includegraphics[width=\linewidth]{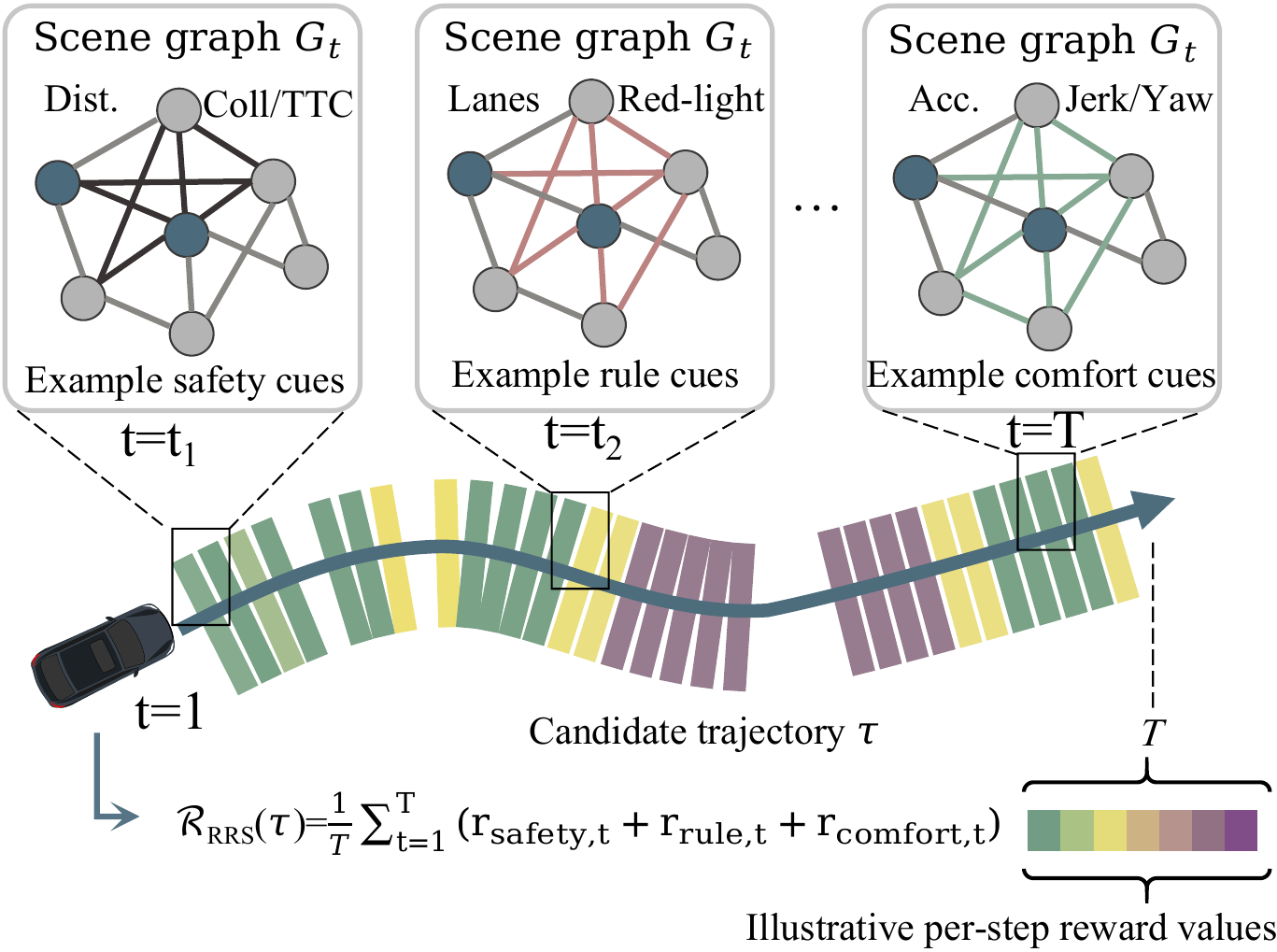}
\caption{Illustration of the relational reward surrogate (RRS). Along the candidate trajectory $\tau$, a time-indexed scene graph $G_t$ is queried at each step to evaluate safety, rule-compliance, and comfort terms. Selected steps are shown for illustration, and the per-step scores are aggregated over the horizon to form the dense trajectory-level reward $\mathcal R_{\mathrm{RRS}}(\tau)$.}
\label{fig:rrs_overview}
\vspace{-0.4cm}
\end{figure}

We introduce a \emph{Relational Reward Surrogate} (RRS) that assigns each ego candidate rollout a dense trajectory-level score by querying a time-indexed ego-centric scene graph, as illustrated in Fig.~\ref{fig:rrs_overview}. It decomposes into safety, rule-compliance, and comfort terms, and provides the trajectory-level supervision used in our unified objective.

\noindent\textbf{Time-expanded surrogate.}
Given a candidate trajectory $\tau=\{(s_t,a_t)\}_{t=1}^T$, where $s_t$ denotes the ego state (including pose and velocity) and $a_t$ the control action at step $t$, we construct at each step a lightweight ego-centric scene graph $G_t=(V_t,E_t)$ that encodes nearby agents, map elements, and traffic signals in a unified relational form. The graph serves as a common interface for querying heterogeneous interactions among agents, lanes, signals, and ego motion. RRS evaluates all three components at every step and aggregates them into a trajectory-level score:
\begin{equation}
\begin{aligned}
\mathcal R_{\mathrm{RRS}}(\tau)
&=\frac{1}{T}\sum_{t=1}^{T}\Big(
r_{\mathrm{safety},t}(s_t,a_t,G_t)
+r_{\mathrm{rule},t}(s_t,a_t,G_t) \\
&\qquad\qquad\qquad\quad
+r_{\mathrm{comfort},t}(s_t,a_t,G_t)
\Big).
\end{aligned}
\label{eq:rrs_total}
\end{equation}
In practice, each reward component is defined as the negative of a nonnegative penalty, such that $\mathcal{R}_{\mathrm{RRS}}(\tau) =-\mathcal{P}_{\mathrm{RRS}}(\tau)\leq 0$; hence, a larger value (closer to zero) indicates fewer violations and better planning quality.

\noindent\textbf{Safety term.}
Let $\mathcal A_t$ be the set of neighboring agent indices in $G_t$. We extract the ego position $p_t^{\text{ego}}$ from $s_t$, and denote the $j$-th neighbor position by $p_t^{(j)}$. We define the closest distance
\begin{equation}
d_t=\min_{j\in\mathcal A_t}\ \|p_t^{\text{ego}}-p_t^{(j)}\|.
\label{eq:rrs_dist}
\end{equation}
If $\mathcal A_t=\varnothing$, we set $d_t=+\infty$. To capture imminent collision risk while avoiding singularities, we further define an inverse time-to-collision (iTTC) metric from relative position and velocity in the ego frame. Let $v_t^{\mathrm{rel}}$ denote the most critical line-of-sight relative speed. Then
\begin{equation}
\mathrm{iTTC}_t=\frac{\max(0,-v_t^{\mathrm{rel}})}{\max(\varepsilon,d_t)},
\label{eq:rrs_ttc}
\end{equation}
where $\varepsilon>0$ avoids division by zero; when $\mathcal A_t=\varnothing$, we set $\mathrm{iTTC}_t=0$. We also include a binary collision indicator $\mathbf{1}_{\mathrm{col},t}$. Using smooth barrier penalties $\phi_{\mathrm{ttc}}(\cdot)$ and $\phi_d(\cdot)$, we define
\begin{equation}
r_{\mathrm{safety},t}
=
-\lambda_{\mathrm{col}}\mathbf{1}_{\mathrm{col},t}
-\lambda_{\mathrm{ttc}}\phi_{\mathrm{ttc}}(\mathrm{iTTC}_t;\mathrm{th}_{\mathrm{ttc}})
-\lambda_d\phi_d(d_t;\mathrm{th}_d),
\label{eq:rrs_safety}
\end{equation}
where $\mathrm{th}_{\mathrm{ttc}}$ and $\mathrm{th}_d$ are thresholds, and $\lambda_{\mathrm{col}},\lambda_{\mathrm{ttc}},\lambda_d>0$ are scalar weights.

\noindent\textbf{Rule compliance and comfort.}
From map-related nodes and edges in $G_t$, we query the lateral deviation $\Delta y_t$, an optional red-light indicator $\mathbf{1}_{\mathrm{red},t}$, and the local speed limit $v_t^{\max}$. Let $v_t$ denote ego speed. We define
\begin{equation}
\begin{aligned}
r_{\mathrm{rule},t}
=&-\lambda_{\mathrm{red}}\mathbf{1}_{\mathrm{red},t}
-\lambda_{\mathrm{lane}}\phi_{\mathrm{lane}}(|\Delta y_t|) \\
&-\lambda_{\mathrm{spd}}\phi_{\mathrm{spd}}\!\left(
\max\!\left(0,\frac{v_t-v_t^{\max}}{\max(\varepsilon_v,v_t^{\max})}\right)
\right),
\end{aligned}
\label{eq:rrs_rule}
\end{equation}
where $\varepsilon_v>0$ is a small constant for numerical stability. For comfort, we penalize acceleration $\mathrm{acc}_t$, jerk $\mathrm{jerk}_t$, and yaw rate $\omega_t$:
\begin{equation}
r_{\mathrm{comfort},t}
=
-\lambda_{\mathrm{acc}}|\mathrm{acc}_t|^2
-\lambda_{\mathrm{jerk}}|\mathrm{jerk}_t|^2
-\lambda_{\omega}\omega_t^2,
\label{eq:rrs_comfort}
\end{equation}
with $\lambda_{\mathrm{acc}},\lambda_{\mathrm{jerk}},\lambda_{\omega}>0$. Here $\mathrm{jerk}_t$ is computed by finite differences of acceleration, with standard boundary handling at the horizon end.
\subsection{Graph-Reward Integrated Diffusion Objective}
\label{ssec:grid_obj}

At each planning step, conditioned on the current scene graph $G$, the diffusion planner samples a rollout
$\tau \sim \pi_\theta(\cdot \mid G)$ and evaluates it using the graph-weighted relational return
\begin{equation}
R_{\mathrm{traj}}(\tau)=w(\tau)\,\mathcal R_{\mathrm{RRS}}(\tau),
\label{eq:return_qsoft}
\end{equation}
where $w(\tau)$ is the graph-aware importance weight defined in Sec.~\ref{ssec:gaiw}, and
$\mathcal R_{\mathrm{RRS}}(\tau)$ is the relational reward surrogate defined in Sec.~\ref{ssec:rrs}.
Our objective is to bias the diffusion policy toward higher-return closed-loop rollouts while preserving the stable denoising structure of the base planner.

\noindent\textbf{Value baseline and advantage estimation.}
We introduce a value baseline $V_\phi(G)$ that predicts the expected return under the current rollout policy. It is trained by on-policy regression:
\begin{equation}
\mathcal L_V(\phi)
=
\mathbb E_{G,\ \tau \sim \pi_\theta(\cdot \mid G)}
\left[
\bigl(V_\phi(G)-R_{\mathrm{traj}}(\tau)\bigr)^2
\right].
\label{eq:critic_loss}
\end{equation}
The corresponding trajectory-level advantage is
\begin{equation}
A(G,\tau)=R_{\mathrm{traj}}(\tau)-V_\phi(G).
\label{eq:advantage}
\end{equation}
Here $A(G,\tau)$ is used only as a stop-gradient improvement signal for reweighting and gating, rather than as a direct policy-gradient estimator.

\noindent\textbf{Advantage-weighted diffusion surrogate.}
Since diffusion planners do not provide a tractable $\log \pi_\theta(\tau\mid G)$, we use the per-sample denoising loss as a surrogate optimization signal. For diffusion step $k$, we perturb the trajectory as
\begin{equation}
z_k=\sqrt{\bar\alpha_k}\,\tau+\sqrt{1-\bar\alpha_k}\,\epsilon,
\qquad
\epsilon\sim\mathcal N(0,I),
\end{equation}
and define the per-sample diffusion loss
\begin{equation}
\ell_{\mathrm{diff}}(\theta;G,\tau,k,\epsilon)
=
\left\|
\epsilon-\epsilon_\theta(z_k,k,G)
\right\|^2.
\label{eq:per_sample_diff}
\end{equation}
To emphasize improved rollouts, we modulate this loss using an advantage-derived weight
\begin{equation}
\gamma(G,\tau)=
\mathrm{clip}\!\left(
\exp\!\left(\frac{\mathrm{stopgrad}(A(G,\tau))}{\lambda_A}\right),
\gamma_{\min},\gamma_{\max}
\right),
\label{eq:adv_weight}
\end{equation}
where $\lambda_A>0$ is a temperature and $\gamma_{\min},\gamma_{\max}>0$ bound the dynamic range. The resulting advantage-weighted diffusion objective is
\begin{equation}
\mathcal L_{\mathrm{AWD}}(\theta)=
\mathbb E_{G,\ \tau\sim\pi_\theta(\cdot\mid G),\ k,\ \epsilon}
\left[
\gamma(G,\tau)\,
\ell_{\mathrm{diff}}(\theta;G,\tau,k,\epsilon)
\right].
\label{eq:awd_loss}
\end{equation}
This preserves the standard diffusion training form while assigning a larger learning weight to trajectories whose graph-conditioned return exceeds the current value baseline.

\noindent\textbf{On-policy self-distillation.}
To further consolidate improved behaviors, we introduce an on-policy self-distillation (OPSD) term. For each scene graph $G$, we sample multiple candidate
trajectories using the same diffusion model with a more expensive
multi-step denoising process and select the highest-return candidate as the teacher $\tau_{\mathrm T}(G)$, which is distilled into a cheaper student prediction $\hat\tau_\theta(G)$. The teacher is treated as a stop-gradient target. We use the advantage only as an improvement gate:
\begin{equation}
\omega(G,\tau)=
\min\!\Bigl(
\mathrm{ReLU}\bigl(\mathrm{stopgrad}(A(G,\tau))\bigr),
\omega_{\max}
\Bigr),
\label{eq:distill_gate}
\end{equation}
where $\omega_{\max}>0$ caps extreme outliers. The OPSD loss is
\begin{equation}
\mathcal L_{\mathrm{OPSD}}(\theta)=
\mathbb E_{G,\ \tau\sim\pi_\theta(\cdot\mid G)}
\left[
\omega(G,\tau)\,
D\!\bigl(\hat\tau_\theta(G),\tau_{\mathrm T}(G)\bigr)
\right],
\label{eq:opsd_loss}
\end{equation}
where $D(\cdot,\cdot)$ is a trajectory distance, e.g., the averaged $\ell_2$ distance over future planning states such as positions or headings.

\noindent\textbf{Unified training objective.}
Finally, we combine the advantage-weighted diffusion loss and the self-distillation term:
\begin{equation}
\mathcal L_{\mathrm{total}}(\theta)=
\mathcal L_{\mathrm{AWD}}(\theta)
+
\lambda_{\mathrm{OPSD}}\,\mathcal L_{\mathrm{OPSD}}(\theta),
\label{eq:total_loss}
\end{equation}
where $\lambda_{\mathrm{OPSD}}>0$ balances distribution shaping and policy consolidation, while $V_\phi$ is updated via Eq.~\eqref{eq:critic_loss}. This unified objective preserves the generative stability of diffusion learning, while graph-weighted relational returns bias training toward trajectories that are safer, more rule-consistent, and more comfortable under closed-loop rollouts.

%% file: sec/4_exp.tex
\section{Experiment}
\label{sec:exp}

\subsection{Datasets and Metrics.}
\label{ssec:datasets_metrics}

\noindent\textbf{nuPlan and interPlan Datasets.}
We evaluate GRID-Planner in closed-loop simulation on two established autonomous driving benchmarks. Following the standard protocol~\cite{cheng2023plantf,zheng2025diffusionplanner}, we report \textbf{nuPlan}~\cite{caesar2021nuplan} results on \textbf{Val14}, \textbf{Test14-random}, and \textbf{Test14-hard}, covering general performance and long-tail robustness. We further evaluate on \textbf{interPlan}~\cite{hallgarten2024vehicle}, which emphasizes interaction-intensive scenarios such as nudging around parked vehicles, unprotected overtaking, and jaywalker encounters.

\noindent\textbf{Evaluation Metrics.}
For \textbf{nuPlan}, we report the official Closed-Loop Score (CLS) under Non-Reactive (NR) and Reactive (R) simulation~\cite{caesar2021nuplan}. CLS aggregates route progress, comfort, and rule compliance while penalizing safety failures such as at-fault collisions. For \textbf{interPlan}, we report the Overall Score and representative scores on \emph{Nudge Around}, \emph{High Traffic Density}, and \emph{Jaywalk}.

\noindent\textbf{Diagnostic Metrics.}
Beyond the official scores, we report \textbf{Collision Rate}, corresponding to $R_{\mathrm{safety}}$; \textbf{Lane Deviation}, a lane/route-consistency proxy corresponding to $R_{\mathrm{rule}}$; \textbf{Jerk}, a smoothness indicator corresponding to $R_{\mathrm{comfort}}$; and \textbf{Latency}, the average planner-side inference time. These metrics are diagnostic only. \textit{Red-light violations are not evaluated separately due to limited traffic-signal state extraction.}

\subsection{Implementation Details}
\label{ssec:impl}

GRID-Planner uses 2 seconds of history to predict an 8-second trajectory at 10 Hz, with 21 historical timestamps including the current frame and 80 future steps. Each waypoint is represented as $(x,y,\cos\theta,\sin\theta)$ in the ego-centric frame. The scene encoder retains up to 32 neighboring agents, 70 lane polylines, 25 route-lane polylines, and 5 static control elements. Training has three stages: (1) base diffusion training for 400 epochs with AdamW (batch size 2048, learning rate $10^{-4}$, weight decay 0.01, and EMA decay 0.999), CFG, and 0.3 neighbor dropout; (2) graph-aware reward fine-tuning with GAIW and RRS for 10 epochs at $5\times10^{-5}$; and (3) online self-distillation for 2 epochs, sampling four candidates per scene and ranking their graph-weighted returns to select the teacher and supervise the student. At inference, the distilled planner uses four update steps with a CFG scale of 1.8. Latency is the mean planner-side \texttt{compute\_trajectory} runtime at batch size 1 on the same hardware, excluding simulator overhead, and is reported only as a relative runtime diagnostic.
\subsection{Main Results}
\label{subsec:main}

\noindent\textbf{nuPlan benchmark.}
Table~\ref{tab:nuplan} reports closed-loop planning results on nuPlan under both non-reactive (NR) and reactive (R) protocols. Among learning-based planners, \textbf{GRID-Planner} achieves the best performance on five of the six splits. Compared with the strong Flow Planner baseline (reproduced locally using its official open-source implementation\footnote{\url{https://github.com/DiffusionAD/Flow-Planner}}), our method improves Val14-R from 79.90 to 82.59 and, more importantly, raises the highly challenging Test14-hard-R score from 66.88 to 71.15, an absolute gain of +4.27. The gains are consistently larger on reactive and hard splits, indicating that \textbf{GRID-Planner} is particularly effective in planning-critical settings, where success depends on resolving dense interactions rather than merely generating plausible nominal trajectories. Similar trends also hold on the non-reactive splits, suggesting that the proposed training strategy improves not only interaction robustness but also overall closed-loop planning quality.

\begin{table}[t]
\centering
\caption{\small Closed-loop planning results of learning-based planners on the nuPlan benchmark under non-reactive (NR) and reactive (R) evaluation protocols. Flow Planner~\citep{tan2025flow} is reproduced locally using its official repository. The best learning-based result in each column is highlighted in bold. $*$ denotes methods that directly leverage the road centerline extracted from the map.}
\vspace{-5pt}
\scriptsize
\setlength{\tabcolsep}{3.2pt}
\renewcommand{\arraystretch}{0.96}
\resizebox{\linewidth}{!}{
\begin{tabular}{lcccccc}
\toprule
\multirow{2}{*}{\textbf{Planner}} 
& \multicolumn{2}{c}{\textbf{Val14}} 
& \multicolumn{2}{c}{\textbf{Test14-hard}} 
& \multicolumn{2}{c}{\textbf{Test14}} \\ 
\cmidrule(lr){2-3} \cmidrule(lr){4-5} \cmidrule(lr){6-7}
& \textbf{NR} & \textbf{R} & \textbf{NR} & \textbf{R} & \textbf{NR} & \textbf{R} \\
\midrule
PDM-Open*~\cite{dauner2023misconceptions} & 53.53 & 54.24 & 33.51 & 35.83 & 52.81 & 57.23 \\
UrbanDriver~\cite{scheel2022urbandriver} & 68.57 & 64.11 & 50.40 & 49.95 & 51.83 & 67.15 \\
GameFormer w/o refine.~\cite{huang2023gameformer} & 13.32 & 8.69 & 7.08 & 6.69 & 11.36 & 9.31 \\
PlanTF~\cite{cheng2023plantf} & 84.27 & 76.95 & 69.70 & 61.61 & 85.62 & 79.58 \\
PLUTO w/o refine.*~\cite{cheng2024pluto} & 88.89 & 78.11 & 70.03 & 59.74 & \textbf{89.90} & 78.62 \\
Flow Planner~\cite{tan2025flow} & 88.21 & 79.90 & 72.17 & 66.88 & 88.29 & 83.07 \\
\midrule
\textbf{GRID-Planner} (\textbf{\textcolor{red}{Ours}}) & \textbf{89.92} & \textbf{82.59} & \textbf{73.86} & \textbf{71.15} & {89.45} & \textbf{85.18} \\
\bottomrule
\end{tabular}}
\label{tab:nuplan}
\vspace{-4mm}
\end{table}

\begin{figure}[t]
    \centering
    \includegraphics[width=1.0\linewidth]{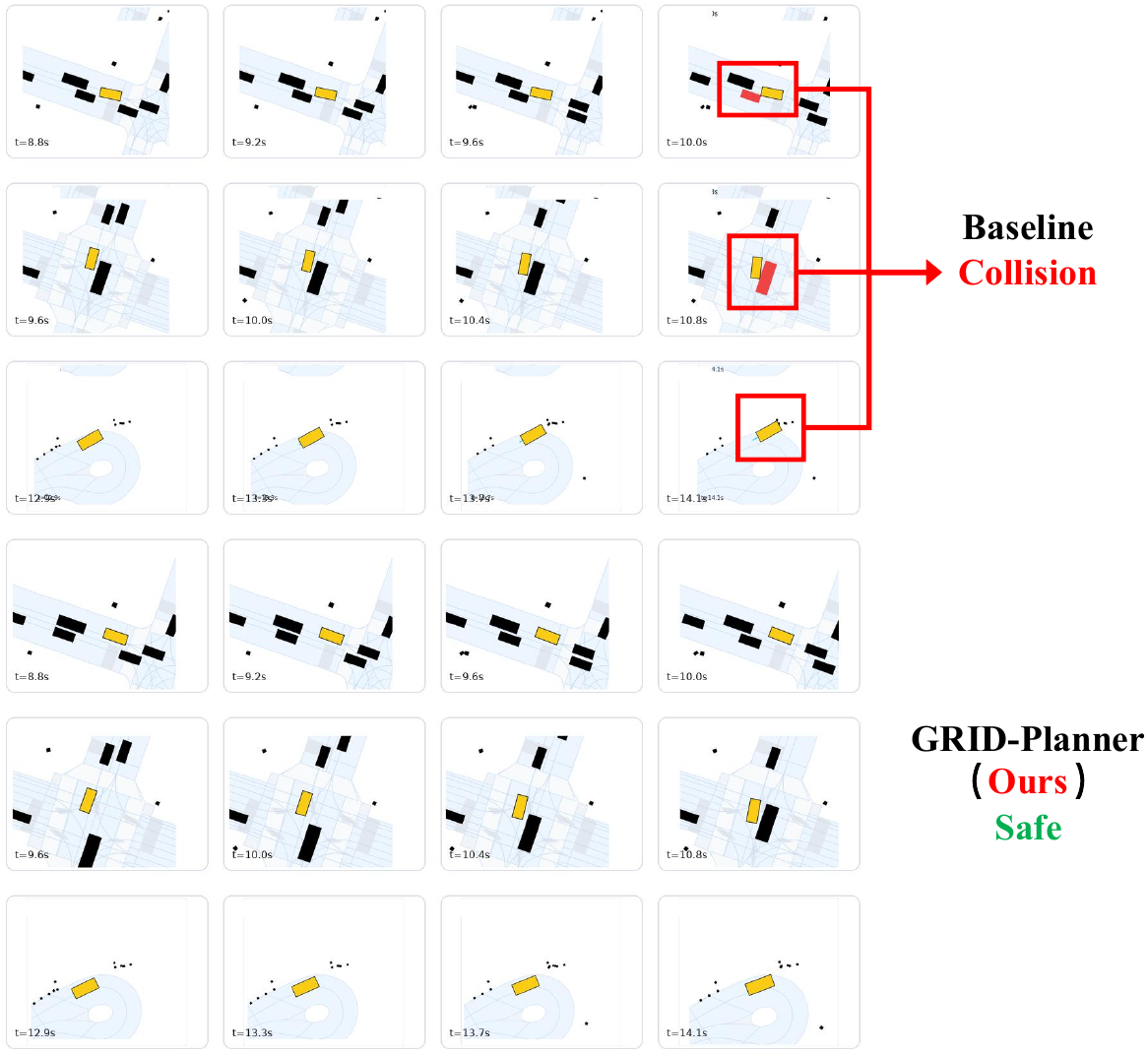}
    \caption{\textbf{Baseline collides while ours remains safe.}
    We visualize closed-loop rollouts in several challenging intersection scenarios. The baseline (top) results in collisions, while our method (bottom) navigates the same scenes. The ego vehicle is highlighted in yellow and collisions are shown in red.}
    \label{fig:vis}
    \vspace{-0.4cm}
\end{figure}

\noindent\textbf{interPlan benchmark.}
Table~\ref{tab:interplan} further evaluates interactive decision making on interPlan, which emphasizes long-tail interactive scenarios beyond standard lane following. \textbf{GRID-Planner} achieves the best overall score of 57.25, outperforming both Diffusion Planner and Flow Planner (52.90 and 54.83, respectively). The improvement is especially evident in spatially constrained and highly uncertain scenarios, including \emph{High Traffic Density} (65.37 vs.\ 56.41 against Flow Planner) and \emph{Jaywalk} (43.20 vs.\ 41.64 against Flow Planner, and 26.20 against Diffusion Planner). Since success in these categories depends on resolving ambiguous right-of-way conflicts rather than nominal path tracking, these results support the role of the proposed relational learning mechanism in shaping safer and more interaction-aware behavior. The gains are not uniform across all categories: in \emph{Nudge Around}, GRID-Planner slightly exceeds Flow Planner (58.77 vs.\ 58.02) but remains below Diffusion Planner (60.48), suggesting that our method is most advantageous in scenarios where relational reasoning and safety--progress trade-offs under dynamic uncertainty dominate the planning difficulty. Qualitative rollouts in Fig.~\ref{fig:vis} further support this observation: in several challenging intersection scenarios, the baseline collides while GRID-Planner remains safe on the same scenes, showing improved closed-loop handling of interactive conflicts.

\begin{table}[t]
\centering
\caption{\small Closed-loop planning results on the interPlan benchmark~\cite{hallgarten2024vehicle}. We report the overall score together with representative scenario-specific scores. The best scores in each column are highlighted in bold.}
\vspace{-2mm}
\scriptsize
\setlength{\tabcolsep}{3pt}
\begin{tabular}{lcccc}
\toprule
\textbf{Planner} 
& \textbf{Overall Score} 
& \textbf{Nudge Around} 
& \textbf{High Traffic Density} 
& \textbf{Jaywalk} \\
\midrule
PlanTF~\cite{cheng2023plantf}            & 47.70 & 49.40 & 58.85 & 33.94 \\
Diffusion Planner~\cite{zheng2025diffusionplanner} & 52.90 & \textbf{60.48} & 49.71 & 26.20 \\
Flow Planner~\cite{tan2025flow}  & 54.83 & 58.02 & 56.41 & 41.64 \\
\midrule
\textbf{GRID-Planner} (\textbf{\textcolor{red}{Ours}}) & \textbf{57.25} & 58.77 & \textbf{65.37} & \textbf{43.20} \\
\bottomrule
\end{tabular}
\vspace{-6mm}
\label{tab:interplan}
\end{table}

\subsection{Ablation Studies}
\label{subsec:ablation}

\noindent\textbf{Main ablation.}
Table~\ref{tab:ablation_main} shows that the major components of GRID-Planner provide complementary gains. Starting from the pure diffusion baseline, progressively introducing graph-aware weighting, relational return shaping, and on-policy optimization steadily improves the official score from 79.90 to \textbf{82.59}. Specifically, adding \textbf{GAIW} raises the score to 80.92 while reducing collision rate, lane deviation, and jerk, indicating that prioritizing planning-critical scenes already strengthens the effective training signal. Adding the graph-weighted trajectory return $R_{\mathrm{traj}}$ further improves the score to 81.31 and yields the strongest comfort and lane-keeping performance at this stage, suggesting better alignment with planning-oriented safety and rule constraints. Building on this, \textbf{on-policy actor--critic} optimization increases the score to 81.92 and substantially reduces collisions, although jerk slightly increases, indicating a temporary comfort--safety trade-off during rollout-based policy improvement. Adding \textbf{OPSD} partially recovers comfort and efficiency while preserving the score gain. The full model achieves the best overall score together with the lowest lane deviation and latency, while maintaining near-best safety and comfort metrics. Overall, these results support the intended role of each component: GAIW emphasizes planning-critical trajectories, {$R_{\mathrm{traj}}$} injects dense relational planning signals, and the on-policy objective with OPSD consolidates these signals into a stronger closed-loop planner.

\begin{table}[t]
\centering
\caption{\textbf{Main ablation} on the full benchmark suite.
Score (\(\uparrow\); nuPlan Val14 Benchmark), safety violations
(\(\downarrow\)), comfort (jerk, \(\downarrow\)), and planning latency
(\(\downarrow\)). All variants use the same training budget and evaluation
protocol.}
\vspace{-2mm}
\scriptsize
\setlength{\tabcolsep}{3.5pt}
\resizebox{\linewidth}{!}{
\begin{tabular}{lccccc}
\toprule
\textbf{Variant} &
\textbf{Score} $\uparrow$ &
\textbf{Coll.} $\downarrow$ &
\textbf{Lane dev.} $\downarrow$ &
\textbf{Jerk} $\downarrow$ &
\textbf{Latency} $\downarrow$ \\
\midrule
Diffusion only ($\mathcal L_{\mathrm{diff}}$)                          & 79.90 & 0.0783 & 0.0268 & 0.1606 & 353.98 \\
+ GAIW ($w(\tau)$)                                                    & 80.92 & 0.0699 & 0.0201 & 0.1302 & 347.56 \\
+ $R_{\mathrm{traj}}$ ($R_{\mathrm{traj}}(\tau)=w(\tau)\mathcal R_{\mathrm{RRS}}(\tau)$) & 81.31 & 0.0604 & 0.0154 & \textbf{0.1153} & 343.72 \\
+ On-policy actor--critic                                             & 81.92 & 0.0503 & 0.0147 & 0.1266 & 345.28 \\
+ On-policy self-distillation                                         & 82.10 & \textbf{0.0501} & 0.0123 & 0.1191 & 341.59 \\
Full model                                                            & \textbf{82.59} & {0.0503} & \textbf{0.0105} & {0.1166} & \textbf{340.53} \\
\bottomrule
\end{tabular}
}
\vspace{-6mm}
\label{tab:ablation_main}
\end{table}

\begin{table}[t]
\centering
\caption{\textbf{GAIW ablation.} Effects of centrality, collision-risk, and stabilization in $w(\tau)$. We report Val14 score (\(\uparrow\)), collision rate (\(\downarrow\)), lane deviation (\(\downarrow\)), jerk (\(\downarrow\)), and latency (\(\downarrow\)). All variants share the same backbone and training budget; only the weighting strategy differs.}
\label{tab:ablation_gaiw}
\vspace{-2mm}
\scriptsize
\setlength{\tabcolsep}{3.2pt}
\renewcommand{\arraystretch}{0.96}
\resizebox{\linewidth}{!}{
\begin{tabular}{lccccc}
\toprule
\textbf{Variant} &
\textbf{Score} $\uparrow$ &
\textbf{Coll.} $\downarrow$ &
\textbf{Lane} $\downarrow$ &
\textbf{Jerk} $\downarrow$ &
\textbf{Latency} $\downarrow$ \\
\midrule
No weighting ($w{=}1$) & {81.61} & 0.0568 & 0.0163 & 0.1440 & 332.14 \\
Centrality only ($w{=}\exp(\alpha_1\mathrm{CentFeat})$) & {81.74} & 0.0559 & 0.0161 & 0.1368 & 332.41 \\
Risk only ($w{=}\exp(\alpha_2\mathrm{CollProb})$) & {81.82} & 0.0546 & 0.0159 & 0.1384 & 332.53 \\
\midrule
Full w/o $w$-clipping & {81.71} & 0.0551 & 0.0159 & 0.1349 & 332.22 \\
Full w/o mean-norm & {81.65} & 0.0547 & 0.0158 & 0.1315 & \textbf{332.07} \\
Full + stabilization & \textbf{82.07} & \textbf{0.0519} & \textbf{0.0154} & \textbf{0.1252} & 332.41 \\
\bottomrule
\end{tabular}}
\vspace{-4mm}
\end{table}

\begin{figure}[t]
\centering
\includegraphics[width=\linewidth]{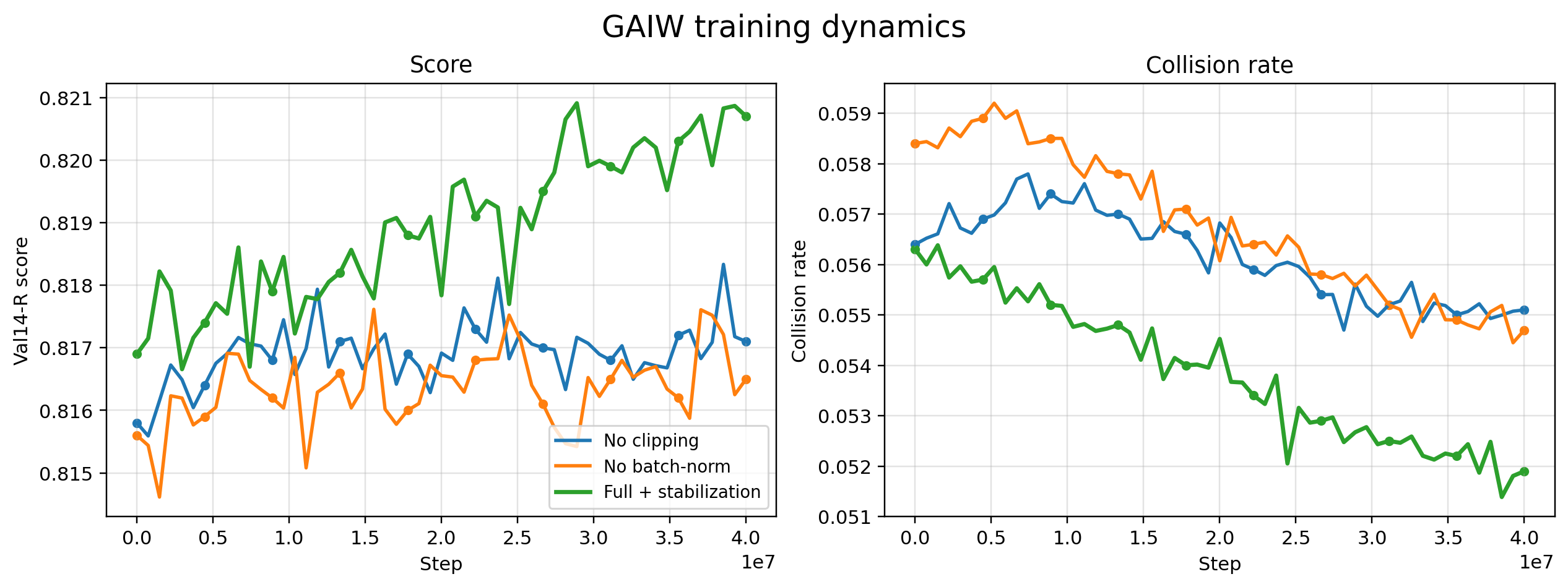}
\caption{\textbf{Training dynamics of GAIW.}
The full GAIW design with clipping and mean-normalization achieves higher
closed-loop score (left) and lower collision rate (right) than
variants without clipping or mean-normalization, indicating improved training stability in planning-critical scenarios.}
\label{fig:gaiw_stability}
\vspace{-0.6cm}
\end{figure}

\noindent\textbf{Ablating the graph-aware weighting mechanism.}
Table~\ref{tab:ablation_gaiw} and Fig.~\ref{fig:gaiw_stability} validate both the effectiveness of GAIW and the necessity of its stabilization design. Compared with no weighting ($w(\tau)=1$), using either graph centrality or collision-risk weighting alone already improves the score ({81.74/81.82 vs.\ 81.61}) while reducing collision rate, lane deviation, and jerk, showing that both structural complexity and explicit risk cues help prioritize planning-critical trajectories. Among the single-factor variants, \textbf{risk-only} weighting yields slightly better score and safety metrics than \textbf{centrality-only}, whereas \textbf{centrality-only} achieves slightly lower jerk, suggesting that the two cues capture complementary aspects of planning difficulty. When both cues are combined, stabilization becomes crucial: removing either $w$-clipping or mean-normalization degrades performance, while \textbf{Full + stabilization} achieves the best overall result, with the highest score (\textbf{{82.07}}) and the lowest collision rate (\textbf{0.0519}), lane deviation (\textbf{0.0154}), and jerk (\textbf{0.1252}). This trend is consistent with Fig.~\ref{fig:gaiw_stability}, where the stabilized variant maintains higher reactive closed-loop score and lower collision rate throughout training than variants without clipping or normalization, indicating improved optimization stability. Latency remains nearly unchanged across all variants, showing that the gain comes from better training signals rather than additional inference cost. Overall, these results support our design intuition: centrality emphasizes structurally complex scenes, collision-risk weighting highlights immediate safety-critical interactions, and clipping plus mean-normalization prevents overly heavy-tailed importance weights from destabilizing optimization.

\noindent\textbf{Ablating the relational reward surrogate (RRS).}
Table~\ref{tab:ablation_rrs} shows that the three components of RRS play distinct yet complementary roles, and that the full surrogate yields the best overall planning behavior. Using the complete reward under the full training pipeline, the planner achieves the highest score (\textbf{82.59}) together with the best collision rate (\textbf{0.0503}), lane deviation (\textbf{0.0105}), and jerk (\textbf{0.1166}), confirming that jointly modeling safety, rule consistency, and comfort is necessary for balanced closed-loop performance. Removing the \textbf{rule} term substantially degrades lane deviation (from 0.0105 to 0.0150) and lowers the overall score, showing that explicit rule-aware shaping is important for route-consistent behavior. Removing the \textbf{comfort} term leaves safety-related metrics relatively stable but worsens jerk (0.1538 vs.\ 0.1166) and increases latency, indicating that the comfort term is important for motion smoothness. The largest degradation occurs when removing the \textbf{safety} term: the score drops markedly to 74.85, while collision rate and lane deviation worsen to 0.0708 and 0.0287, respectively, demonstrating that safety remains the dominant constraint in closed-loop planning. The single-term variants further support this decomposition: \textbf{safety only} preserves relatively better collision performance, \textbf{rule only} better preserves lane consistency, and \textbf{comfort only} achieves the smoothest trajectories with the lowest jerk but performs poorly on overall planning quality and safety. Overall, these results match the intended design of RRS: the safety term anchors collision avoidance, the rule term enforces scene-consistent behavioral constraints, and the comfort term regularizes motion smoothness; only their combination provides a dense and well-balanced planning signal aligned with the full closed-loop objective.

\begin{table}[t]
\centering
\caption{\textbf{RRS ablation.} Removing individual components of
$\mathcal R_{\mathrm{RRS}}(\tau)$ reveals their targeted impact on safety,
rule consistency, and comfort. Score (\(\uparrow\); nuPlan Val14 Benchmark),
safety violations (\(\downarrow\)), comfort (jerk, \(\downarrow\)), and
planning latency (\(\downarrow\)) are reported. All variants use the full
training pipeline and differ only in the reward surrogate.}
\vspace{-2mm}
\scriptsize
\setlength{\tabcolsep}{3.5pt}
\resizebox{\linewidth}{!}{
\begin{tabular}{lccccc}
\toprule
\textbf{Variant} &
\textbf{Score} $\uparrow$ &
\textbf{Coll.} $\downarrow$ &
\textbf{Lane dev.} $\downarrow$ &
\textbf{Jerk} $\downarrow$ &
\textbf{Latency} $\downarrow$ \\
\midrule

Full RRS ($R_{\mathrm{safety}}+R_{\mathrm{rule}}+R_{\mathrm{comfort}}$)
& \textbf{82.59}
& \textbf{0.0503}
& \textbf{0.0105}
& \textbf{0.1166}
& \textbf{340.53}
\\

w/o rule term ($R_{\mathrm{safety}}+R_{\mathrm{comfort}}$)
& 80.12
& 0.0568
& 0.0150
& 0.1325
& 344.87
\\

w/o comfort term ($R_{\mathrm{safety}}+R_{\mathrm{rule}}$)
& 81.05
& 0.0527
& 0.0116
& 0.1538
& 345.92
\\

w/o safety term ($R_{\mathrm{rule}}+R_{\mathrm{comfort}}$)
& 74.85
& 0.0708
& 0.0287
& 0.1786
& 348.21
\\

\midrule

Safety only ($R_{\mathrm{safety}}$)
& 79.62
& 0.0555
& 0.0248
& 0.1451
& 342.96
\\

Rule only ($R_{\mathrm{rule}}$)
& 78.91
& 0.0613
& 0.0158
& 0.1546
& 340.82
\\

Comfort only ($R_{\mathrm{comfort}}$)
& 72.95
& 0.0739
& 0.0538
& \textbf{0.1288}
& \textbf{333.41}
\\

\bottomrule
\end{tabular}
}
\vspace{-4mm}
\label{tab:ablation_rrs}
\end{table}

\begin{table}[t]
\centering
\caption{\textbf{Unified on-policy objective ablation.} Each row removes one
ingredient from Eq.~\ref{eq:total_loss} while keeping the graph-aware return
$R(\tau,G)$, rollout budget, and evaluation protocol fixed. We report Score 
(\(\uparrow\); nuPlan Val14 Benchmark), together with collision rate 
(\(\downarrow\)), lane deviation (\(\downarrow\)), comfort measured by jerk 
(\(\downarrow\)), and latency (\(\downarrow\)). The last two rows 
ablate OPSD-specific choices (advantage gating and teacher selection).}
\vspace{-2mm}
\scriptsize
\setlength{\tabcolsep}{3.5pt}
\resizebox{\linewidth}{!}{
\begin{tabular}{lccccc}
\toprule
\textbf{Variant} &
\textbf{Score} $\uparrow$ &
\textbf{Coll.} $\downarrow$ &
\textbf{Lane dev.} $\downarrow$ &
\textbf{Jerk} $\downarrow$ &
\textbf{Latency} $\downarrow$ \\
\midrule

Offline diffusion only ($\mathcal L_{\mathrm{diff}}$) 
& {79.90} 
& 0.0783 
& 0.0268 
& 0.1606 
& 353.98 \\

Unified objective (Eq.~\ref{eq:total_loss}) 
& \textbf{{82.59}} 
& \textbf{0.0503} 
& \textbf{{0.0105}} 
& \textbf{{0.1166}} 
& \textbf{{340.53}} \\

\midrule

w/o critic baseline ($V_\phi(G)=0$) 
& {80.03} 
& 0.0754 
& 0.0264 
& 0.1558 
& 345.26 \\

w/o advantage reweighting 
& {80.76} 
& \underline{0.0564} 
& 0.0220 
& 0.1315 
& \underline{344.41} \\

w/o OPSD term 
& {80.29} 
& 0.0699 
& 0.0235 
& 0.1321 
& 344.46 \\

\midrule

w/o advantage gate ($\omega(G)=1$) 
& \underline{{80.89}} 
& 0.0559 
& \underline{0.0176} 
& \underline{0.1272} 
& \textbf{340.53} \\

w/ random teacher 
& {79.57} 
& 0.0894 
& 0.0397 
& 0.1753 
& 372.02 \\

\bottomrule
\end{tabular}
}
\vspace{-4mm}
\label{tab:ablation_grid_obj}
\end{table}

\begin{figure}[t]
\centering
\includegraphics[width=\linewidth]{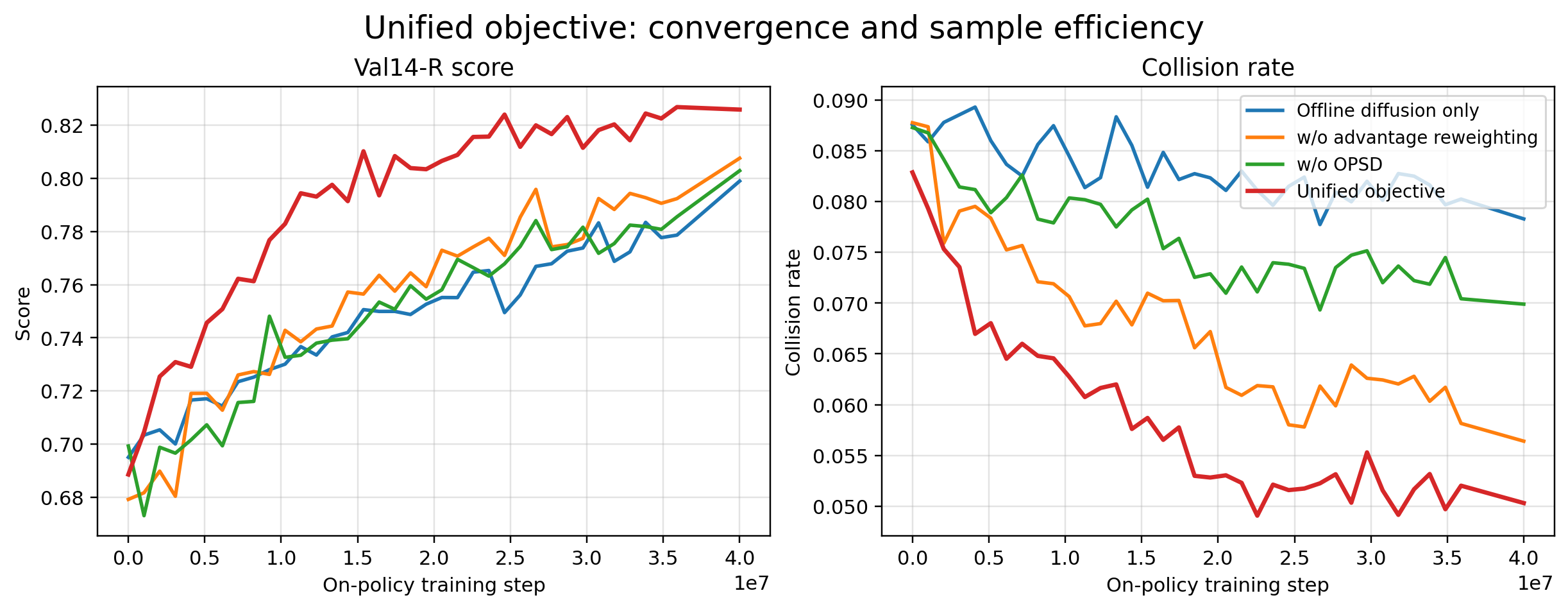}
\caption{\textbf{Training dynamics of the unified on-policy objective.}
The full graph-reward integrated objective shows faster convergence and lower
collision rate than offline diffusion and partial ablations, demonstrating
better sample efficiency and more stable closed-loop adaptation.}
\label{fig:grid_obj_sample_eff}
\vspace{-0.6cm}
\end{figure}

\noindent\textbf{Ablating the graph-reward integrated objective.}
Table~\ref{tab:ablation_grid_obj} and Fig.~\ref{fig:grid_obj_sample_eff} show that the full on-policy objective is essential for achieving the best overall closed-loop performance. Compared with offline diffusion only, the unified objective improves the score from {79.90} to \textbf{{82.59}}, while substantially reducing collision rate (0.0783 $\rightarrow$ \textbf{0.0503}), lane deviation (0.0268 $\rightarrow$ \textbf{{0.0105}}), and jerk (0.1606 $\rightarrow$ \textbf{{0.1166}}), suggesting that graph-reward shaping is most effective when coupled with on-policy optimization rather than used as a purely offline surrogate. Removing the \textbf{critic baseline} yields only marginal gain over the offline baseline (80.03 vs.\ 79.90), indicating that value estimation is important for producing a usable low-variance improvement signal. Removing \textbf{advantage reweighting} or the \textbf{OPSD} term also degrades performance, with score dropping to 80.76 and 80.29, respectively, together with worse safety and tracking metrics, showing that reweighting helps focus learning on better rollouts while OPSD helps consolidate these improvements into the policy. The OPSD-specific ablations further clarify this effect: removing the \textbf{advantage gate} preserves relatively competitive collision, lane, jerk, and latency metrics, but still lowers the score to 80.89, suggesting that ungated distillation weakens overall policy improvement; using a \textbf{random teacher} causes the largest degradation (79.57 score, 0.0894 collision rate, 0.1753 jerk), demonstrating that teacher quality is crucial for successful consolidation. This trend is also reflected in Fig.~\ref{fig:grid_obj_sample_eff}, where the full objective exhibits faster convergence and consistently lower collision rate during training. Overall, the results support the intended role of each component: the critic baseline stabilizes advantage estimation, advantage reweighting emphasizes better trajectories, and OPSD transfers these gains into a stronger closed-loop planner.